\documentclass[12pt]{iopart}
\usepackage{harvard}
\usepackage[table]{xcolor}
\usepackage{graphicx}
\usepackage{multirow}
\usepackage{arydshln}
\usepackage{lineno}
\usepackage[colorlinks=true,linkcolor=blue]{hyperref}

\begin{document}

\title[Learning-based estimation of dielectric properties for radio-frequency]{Learning-based estimation of dielectric properties and tissue density in head models for personalized radio-frequency dosimetry\footnote{\it \color{blue}This work is submitted to Phys. Med. Biol.}}

\author{Essam A Rashed$^{1,2}$, Yinliang Diao$^{1,3}$, Akimasa Hirata$^{1,4}$}

\address{$^1$Department of Electrical and Mechanical Engineering, Nagoya Institute of Technology, Nagoya 466-8555, Japan}
\address{$^2$Department of Mathematics, Faculty of Science, Suez Canal University, Ismailia 41522, Egypt}
\address{$^3$College of Electronic Engineering, South China Agricultural University, Guangzhou 510642, China}
\address{$^4$Center of Biomedical Physics and Information Technology, Nagoya Institute of Technology, Nagoya 466-8555, Japan}
\ead{essam.rashed@nitech.ac.jp}

\vspace{10pt}


\begin{abstract}
Radio-frequency dosimetry is an important process in assessments for human exposure safety and for compliance of related products. Recently, computational human models generated from medical images have often been used for such assessment, especially to consider the inter-subject variability. However, a common procedure to develop personalized models is time consuming because it involves excessive segmentation of several components that represent different biological tissues, which is a major obstacle in the inter-subject variability assessment of radiation safety. Deep learning methods have been shown to be a powerful approach for pattern recognition and signal analysis. Convolutional neural networks with deep architecture are proven robust for feature extraction and image mapping in several biomedical applications. In this study, we develop a learning-based approach for fast and accurate estimation of the dielectric properties and density of tissues directly from magnetic resonance images in a single shot. The smooth distribution of the dielectric properties in head models, which is realized using a process without tissue segmentation, improves the smoothness of the specific absorption rate (SAR) distribution compared with that in the commonly used procedure. The estimated SAR distributions, as well as that averaged over 10-g of tissue in a cubic shape, are found to be highly consistent with those computed using the conventional methods that employ segmentation.
\end{abstract}

\vspace{2pc}

\citationmode{abbr}

\section{Introduction}

In the last few decades, many anatomical human models have been developed, especially for computational dosimetry~\cite{Nagaoka2004PMB,Dimbylow1997PMB,Christ2005BEM,Lee2015TEMC}, to quantify the induced physical quantities in a target tissue for electromagnetic field exposures. Such models have been developed from different types of medical images, e.g., magnetic resonance imaging (MRI), computed tomography (CT), and other modalities. Typically, the models are segmented into several tissue types for the head and approximately 50 or more for the whole body. In general, identical physical constants are assigned to each tissue. For example, in human-head studies, the standard process is first to segment major head tissues. Then, tissue-based constants from literature are uniformly assigned to each segmented tissues. This assumption is unrealistic because even within the same tissue, the dielectric properties can vary based on different parameters such as water content~\cite{Michel2017MRM}, sodium concentration~\cite{Liao2019MRM}, and anatomical structure. This biological phenomenon is notable especially in the tissue border regions where the tissue physical properties have large differences~\cite{Gurler2017MRM}. Moreover, segmentation is usually conducted using different approaches, which may lead to high variability in dosimetry studies. Accurate segmentation is usually considered as a computationally expensive process that limits its feasibility for clinical use.

To avoid potential errors caused by segmentation faults, several methods have been proposed to directly estimate the tissue dielectric properties based on anatomical images~\citeaffixed{Serralles2020,Hampe2018MRM,Ropella2017MRM,Elsaid2017TMI,Tuch2001PNAS}{e.g.}. The water content calculated from T1-weighted MR scans is modeled using a monotonic function to estimate the conductivity of major brain tissues~\cite{Michel2017MRM}. However, such methods utilize data that are strictly limited to the brain only. Magnetic resonance electrical impedance tomography (MREIT) has been proposed as a useful approach to estimate brain conductivity~\cite{Chauhan2018TMI,Kwon2016TBE1,Kwon2014PMB}. A recent review of the methods to estimate the electrical conductivity based on MR scans was presented in~\cite{Liu2017TBE}. In addition to the electrical conductivity, relative permittivity is required in radio-frequency dosimetry. Moreover, tissue density is required for the specific absorption rate (SAR) computations.

Learning-based techniques are now the state of the art in several pattern recognition and data labeling problems~\cite{Bengio2013TPAMI}. Interesting achievements can be observed, and significant improvement has been reported in the past few years compared with the conventional methods. The most important key for the success of the learning-based approaches is the architecture depth that enables better extraction of important features without the need for prior engineering or customized regularizers. In particular, convolutional neural networks (CNNs) are now the leading tool for image processing and recognition. 

\begin{figure}
\centering
\includegraphics[width=\textwidth]{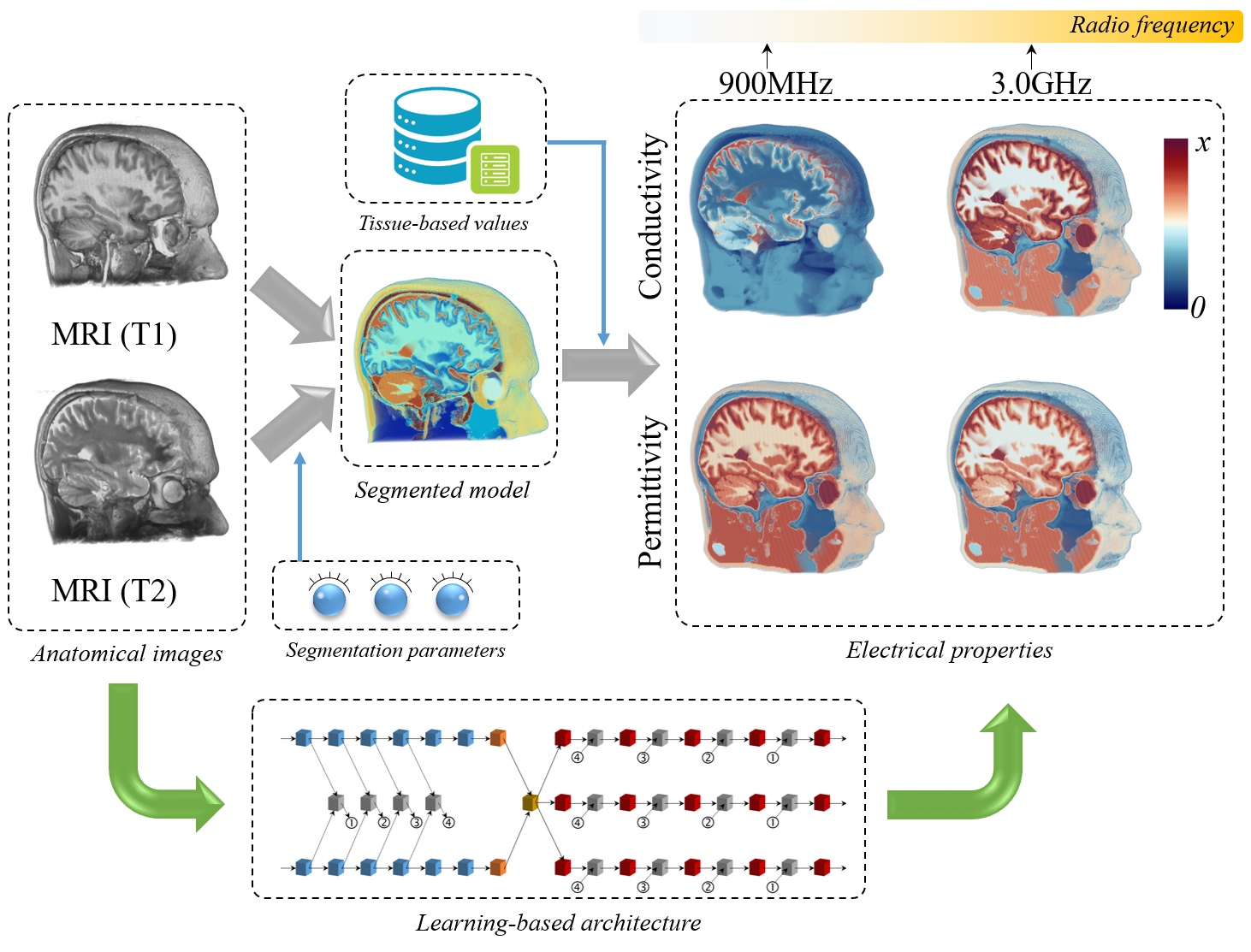}
\caption{The current standard pipeline based on tissue segmentation is shown by gray arrows. The anatomical images are segmented to generate the annotated head model. Then, the tissue-based electrical properties are assigned to each tissue based on the desired frequency. Finally, different conductivity and relative permittivity models are generated using the tissue-based uniform values. The green arrows indicate the contribution of this study in which several parameter adjustments can be avoided using the learning-based architecture. The values are $x$ = 3.0 S/m and $x$ = 70 in the conductivity and permittivity maps, respectively.}
\label{fig01}
\end{figure}

For human exposure to non-ionizing electromagnetic fields below the frequency of 300 GHz, two regimes exist from the viewpoint of electromagnetics. The criterion for transition is when the magneto- or electro-quasi-static approximation is applicable. The transition frequency is approximately 100 kHz-10 MHz, although it depends on the source type and dimension~\cite{Hirata2013PMB,Jin2012PMB,Hand2008PMB}. For exposure to lower frequencies ($<$ 10 MHz) where the quasi-static approximation is applicable, non-invasive brain stimulation is one of the emerging medical applications. Such approaches may be used to avoid the significant inconsistency caused by the inter- and intra-subject variability~\cite{Rashed2019Pulse,Laakso2015BS,Lopezalonso2014BS}. In a previous study, we succeeded in applying deep learning to estimate the electrical conductivity based on MRI in low-frequency medical applications such as transcranial magnetic stimulation (TMS)~\cite{Rashed2020TMI}. The network architecture, which is known as CondNet, was able to provide high-quality estimation of a brain-induced electric field without tissue segmentation.

At frequencies higher than 10 MHz, personalized head models are used for radio-frequency hyperthermia \citeaffixed{DAndrea2007}{e.g.}. In this regime, relative permittivity is also needed for full-wave analysis. In addition, the standardization of human protection from electromagnetic fields considers the inter-subject variability, which in general considered in the reduction factor of the limit, and thus, the dosimetric evaluation of different human models has become essential~\cite{Liu2019IJERPH,Susnjara2018SpliTech,Li2018EUCAP}.

In this study, we first extended CondNet architecture, which was used in low-frequency dosimetry, to the radio-frequency dosimetry studies. The primary difference between low-frequency and radio-frequency is the necessity of relative permittivity and tissues density information. Therefore, CondNet is redesigned to fit with the requirements of radio-frequency dosimetry. Then, the validity of the proposal is evaluated for SAR evaluation for a dipole antenna in GHz bands, which is a canonical exposure scenario from handset antennas.

\begin{table}
\centering
\footnotesize
\caption{Human head tissue conductivity ($\sigma$) [S/m], relative permittivity ($\varepsilon$), and mass density ($\rho$)~[kg/m$^3$] values at different radio-frequencies (Gabriel et al. 1996).}
\setlength{\tabcolsep}{3pt}
\begin{tabular}{ c|l |c|c |c  |  c| c  |  c| c}
\hline
\multirow{2}{*}{\#}&\multirow{2}{*}{\bf Tissue}	& $\rho$  &\multicolumn{2}{c|}{ 900 MHz}  & \multicolumn{2}{c|}{ 1.8 GHz} & \multicolumn{2}{c}{ 3.0 GHz}	\\
\cline{4-9} 
&& (kg/m$^3$)&$\sigma$ & $\varepsilon$ &  $\sigma$ & $\varepsilon$&  $\sigma$ & $\varepsilon$\\
\hline
\hline
~1&Blood                & 1050 & 1.54 & 61.36 & 2.04 & 59.37 & 3.05 & 57.35 \\
~2&Bone (Cancellous)    & 1178 & 0.34 & 20.79 & 0.59 & 19.34 & 1.01 & 17.94 \\
~3&Bone (Cortical)      & 1908 & 0.14 & 12.45 & 0.28 & 11.78 & 0.51 & 11.07 \\ 
~4&Brain (Grey Matter)  & 1145 & 0.94 & 52.73 & 1.39 & 50.08 & 2.22 & 48.05 \\
~5&Brain (White Matter) & 1041 & 0.59 & 38.89 & 0.91 & 37.01 & 1.51 & 35.54 \\
~6&Cerebellum           & 1045 & 1.26 & 49.44 & 1.71 & 46.11 & 2.48 & 43.90 \\
~7&Cerebro Spinal Fluid & 1007 & 2.41 & 68.64 & 2.92 & 67.20 & 4.01 & 65.39 \\
~8&Dura                 & 1174 & 0.96 & 44.43 & 1.32 & 42.89 & 2.01 & 41.34 \\
~9&Fat                  & ~911 & 0.05 & ~5.46 & 0.08 & ~5.35 & 0.13 & ~5.22 \\
10&Mucous tissue        & 1102 & 0.84 & 46.08 & 1.23 & 43.85 & 1.95 & 42.11 \\
11&Muscle               & 1090 & 0.94 & 55.03 & 1.34 & 53.55 & 2.14 & 52.06 \\
12&Skin                 & 1109 & 0.87 & 41.41 & 1.18 & 38.87 & 1.74 & 37.45 \\
13&Vitreous Humor       & 1005 & 1.64 & 68.90 & 2.03 & 68.57 & 2.96 & 67.82 \\
\hline
\end{tabular}
\label{tab01}
\end{table}

\begin{figure}
\centering
\includegraphics[width=\textwidth]{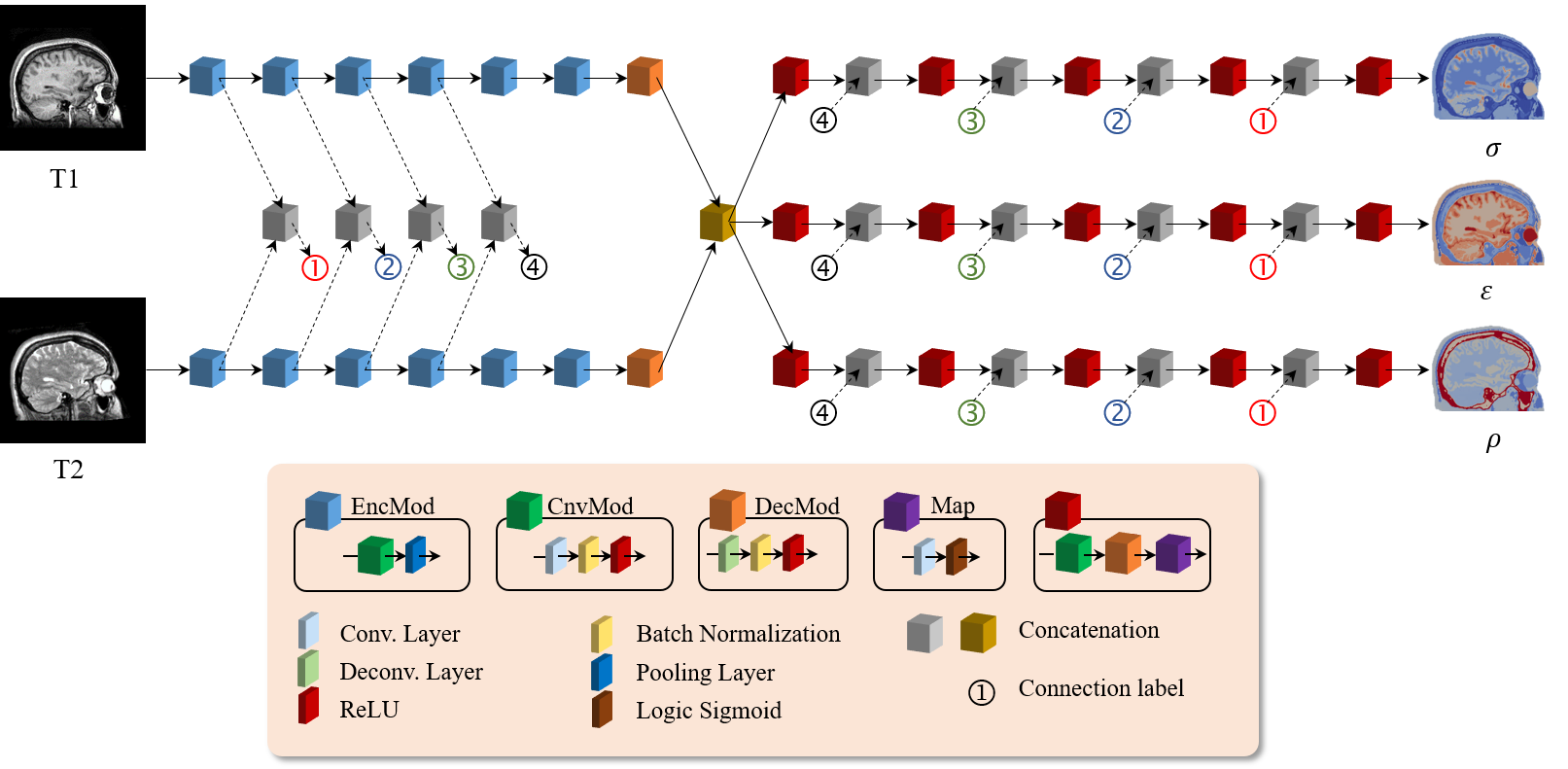}
\caption{Design of the learning-based architecture and label key. The inputs are T1- and T2-weighted MRI, and the outputs are the conductivity ($\sigma$), relative permittivity ($\varepsilon$), and density ($\rho$) maps. The dashed lines represent the skip connections, and the architecture feature size is listed in Table~\ref{tab02}.}
\label{fig02}
\end{figure}

\section{Materials and Methods}

Learning-based architecture is designed to estimate the dielectric properties, i.e., relative permittivity, in addition to the electrical conductivity, required for radio-frequency dosimetry using anatomical information from T1- and T2-weighted MRI. Moreover, it estimates the tissue mass density for the computation of SAR, which is defined as the power absorption per unit mass. A schematic demonstration of the current standard pipeline for computation of the head models for SAR calculations and the proposed approach is shown in figure~\ref{fig01}.


\subsection{Anatomical dataset and preprocessing}

In this study, a set of anatomical images corresponding to 18 subjects from the NAMIC: Brain Mutlimodality dataset\footnote{http://hdl.handle.net/1926/1687} are used. Each subject is assigned to T1- and T2-weighted MR scans with a volume size of 256$^3$ voxels and resolution of 1.0 mm$^3$. Each subject is segmented into 13 different tissues using a semi-automatic method~\cite{Laakso2015BS}, as listed in Table~\ref{tab01}. The segmented models are assigned with uniform values of conductivity ($\sigma$), relative permittivity ($\varepsilon$), and density ($\rho$), as listed in Table~\ref{tab01}. The anatomical magnetic resonance images are normalized such that each subject has a zero mean and unit variance, followed by scaling in the range of $[0,1]$. The conductivity, relative permittivity, and mass-density values are normalized using the following equations:

\begin{equation}
\tilde{\sigma}^r=\frac{(1-\tau_{\sigma})}{\max_n (\sigma_n^r)}\sigma^r,~~\tilde{\varepsilon}^r=\frac{(1-\tau_{\varepsilon})}{\max_n (\varepsilon_n^r)}(\varepsilon^r-1) ,~~\tilde{\rho}=\frac{(1-\tau_{\rho})}{\max_n (\rho_n)}\rho,
\label{eq01}
\end{equation}
where $\sigma^r_n$ and $\varepsilon^r_n$ are the uniform conductivity and permittivity measured for tissue $n$ at frequency $r$, respectively. $\tau_{\sigma}$,$\tau_{\varepsilon}$, and $\tau_{\rho}$ are scaling parameters (here, $\tau_x=0.1$, $x\equiv \sigma$, $\varepsilon$, and $\rho$).

\begin{table*}
\centering
\footnotesize
\caption{Details of learning architecture shown in figure~\ref{fig02}.}
\begin{tabular}{|l|ll|c|}
\hline 
{\bf Module} & {\bf Layer} & {\bf Output size} &    {\bf Label}\\
\hline \hline
Input$_u$ & & \multirow{2}{*}{256$\times$256} &\\
$u:1 \rightarrow 2$ &&   &\\

\hline
EncMod$_{u,i}$ & Convolution & $2^{(i+1)} \times [2^{(8-i)}]^2 $ &  \multirow{3}{*}{\includegraphics[width=.5cm]{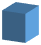}}\\
$u:1 \rightarrow 2$& BN \& ReLU & $2^{(i+1)} \times [2^{(8-i)}]^2 $ &    \\
$i:1 \rightarrow 6$& Pooling (Max) & $2^{(i+1)} \times[2^{(7-i)}]^2 $ & \\

\hline
DecMod$_{u}$ & Deconvolution & 64$\times$8$\times$8 &  \multirow{2}{*}{\includegraphics[width=.5cm]{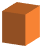}} \\
$u:1 \rightarrow 2$ & BN \& ReLU & 64$\times$8$\times$8   & \\

\hline
\multirow{2}{*}{Hub} & \multirow{2}{*}{Concatenation} & \multirow{2}{*}{2$\times$64$\times$8$\times$8} &\multirow{2}{*}{\includegraphics[width=.5cm]{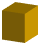}}\\
 &  &   &\\

\hline
CnvMod$_{v,i}$ & Convolution & $2^{(i+2)} \times [2^{(8-i)}]^2 $ & \multirow{3}{*}{\includegraphics[width=.5cm]{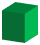}}\\
$v:1 \rightarrow V$ & BN \& ReLU & $2^{(i+2)} \times [2^{(8-i)}]^2 $   & \\
$i:5 \rightarrow 1$ & & &   \\

\hline
DecMod$_{v,i}$ & Deconvolution & $2^{(i+1)} \times [2^{(9-i)}]^2$ &  \multirow{3}{*}{\includegraphics[width=.5cm]{Table2Dec}} \\
$v:1 \rightarrow 3$ & BN \& ReLU &  $2^{(i+1)} \times [2^{(9-i)}]^2$  & \\
$i:5 \rightarrow 1$ &  &        &\\

\hline
Map$_{v,i}$ & Convolution & $2^i \times [2^{(9-i)}]^2$ &\multirow{4}{*}{\includegraphics[width=.5cm]{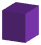}}\\
$v=1 \rightarrow 3$  & Sigmoid (Log) &$ \left\{\begin{array}{ll}
2^i \times [2^{(9-i)}]^2 & i > 1 \\
{[2^8]}^2 & i=1  \end{array} \right.$ &\\
$i:5 \rightarrow 1$ & &  &   \\

\hline
Concat$_{v,i}$ &  &  & \multirow{3}{*}{\includegraphics[width=.5cm]{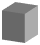}}\\
$v:1 \rightarrow 3$ & Concatenation &  $3\times 2^{(i+2)} \times [2^{(8-i)}]^2$ & \\
$i=4 \rightarrow 1$ &  &  & \\

\hline
Output$_v$ & & \multirow{2}{*}{ 256$\times$256} &\\
$v=1 \rightarrow 3$ & &  &\\
\hline
\end{tabular}
\label{tab02}
\end{table*}


\subsection{Learning architecture}

The objective of the learning architecture is to map the anatomical images ($M^{T1}$ and $M^{T2}$) to the dielectric properties and tissue density ($\sigma^r$, $\varepsilon^r$, and $\rho$) at frequency $r$. Therefore, the learning-architecture parameters are optimized to fit the training data by solving the following problem:

\begin{equation}
\min_\Omega f(M,\theta^r|\Psi),
\end{equation}
where $M=\{M^{T1},M^{T2}\}$, $\theta^r=\{\tilde{\sigma}^r, \tilde{\varepsilon}^r, \tilde{\rho}\}$. $f$ is the loss function, and $\Psi$ represents the network parameters. The deep-learning architecture CondNet presented in \citeasnoun{Rashed2020TMI} is extended to fit with the problem considered in this study. The architecture shown in figure~\ref{fig02}, which maps a dual input of T1- and T2-weighted MRI slices to a triple output, demonstrates the conductivity, relative permittivity, and tissue-density distributions. The main components of this learning architecture are the convolution and deconvolution operations, which concurrently operate to map the gray-scale values in MRI to the corresponding physical properties of the tissue. Encoders with several convolution operations are designed to independently learn the anatomical features, whereas decoders are developed to map these features to different tissue properties in a parallel manner. This strategy proves to be useful in improving the network learning \cite{Rashed2019neuroimage,Rashed2020TMI}. The size of the learned features at each network layer is listed in Table~\ref{tab02}. The estimated normalized physical properties are computed as follows:

\begin{equation}
\{\sigma_k^r, \varepsilon_k^r, \rho_k\}=\textnormal{CondNet}_r (M_k^{T1},M_k^{T2}), ~\forall k
\end{equation}
where $M^{T1}$ and $M^{T2}$ are the normalized T1- and T2-weighted MRI volumes, respectively, and subscript $k$ denotes the slice number. The learning process is considered using slices extracted from the axial, sagittal, and coronal orientations. Therefore, the estimated physical properties are averaged as follows:

\begin{equation*}
\sigma^r_*=\frac{1}{3} (\sigma^r_a+\sigma^r_s+\sigma^r_c), 
\end{equation*}
\begin{equation*}
\varepsilon^r_*=\frac{1}{3} (\varepsilon^r_a+\varepsilon^r_s+\varepsilon^r_c),
\end{equation*}
\begin{equation}
\rho_*=\frac{1}{3} (\rho_a+\rho_s+\rho_c),
\end{equation}
where $\sigma_a$, $\sigma_s$, and $\sigma_c$ are the conductivity volume maps estimated using the axial, sagittal, and coronal slices, respectively. In the same manner, $\varepsilon_*$ and $\rho_*$ are computed. The learning-based conductivity, relative permittivity, and tissue density maps are computed by re-scaling using the following equations:

\begin{equation}
\hat{\sigma}^r=\frac{\max_n(\sigma^r_n)}{1-\tau_{\sigma}}\sigma^r_*, ~~ \hat{\varepsilon}^r=1+\frac{\max_n(\varepsilon^r_n)}{1-\tau_{\varepsilon}}\varepsilon^r_*, ~~\hat{\rho}=\frac{\max_n(\rho_n)}{1-\tau_{\rho}}\rho_*.
\end{equation}

\begin{figure}
\centering
\includegraphics[width=0.3\textwidth]{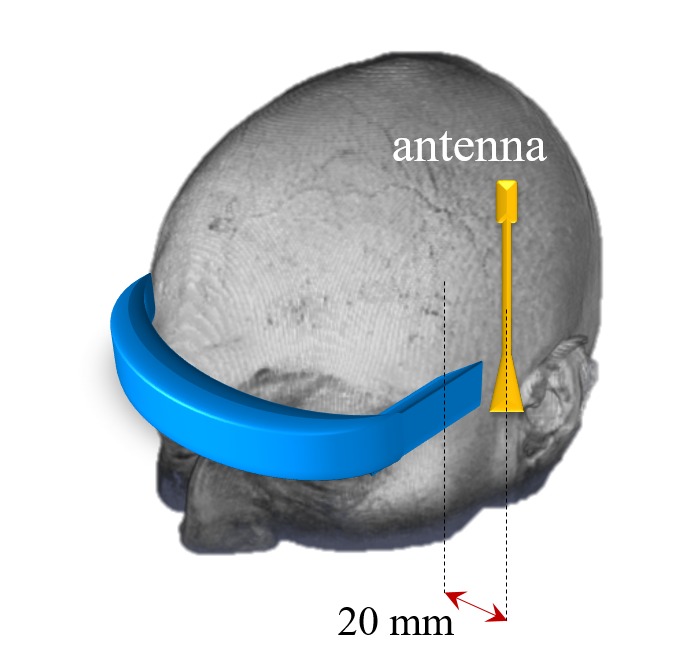}
\caption{Demonstration of the simulated antenna position at the left temporal lobe.}
\label{fig03}
\end{figure}


\subsection{Radio-frequency exposure from a dipole antenna}

We consider a scenario where the head is exposed to a dipole antenna at radio frequencies. This setup is a canonical one considering the exposure from wireless communication devices, e.g., mobile phones. The radiation source is a half-wavelength dipole antenna located at 20 mm from the scalp close to the temporal lobe, as shown in figure~\ref{fig03}, with an emitted power of 1 W. The finite-difference time-domain (FDTD) method \cite{Taflove2005} is adopted to solve the electromagnetic problem. The dipole antenna is orientated parallel to the longitudinal axis of the body. The antenna lengths are 15.7, 7.9, and 4.7 cm at 0.9, 1.8, and 3.0 GHz, respectively. A 10-layer convolutional perfectly matched layer (CPML) \cite{Roden2000MOTL} is adopted to truncate the simulation domain. The total simulation domain contains 290$^3$ voxels. The SAR is obtained by the following:

\begin{equation}
SAR=\frac{\sigma(t)}{\rho(t)} |E(t)|^2,
\end{equation}
where $|E(t)|$ is the rms value of the calculated electric field at $t(x,y,z)$ inside the head model.

Canonical exposure scenario including the antenna was based on the international exposure standardization, in which a relatively simple scenario  is often considered \cite{ICNIRP2019,IEEEC9512019}. The allocated frequency bands for existing mobile communication services are mainly in the range from 900~MHz to 3~GHz. Also most common hand-held mobile devices are designed to be operated near the head sides, especially when the antenna is transmitting EM signal. The selection of the frequencies in this study is also based on the consideration of the penetration depth of EM wave. The penetration depth in high-water-content human tissue at 3.0 GHz is approximately 1.6 cm, and decreases with the increased frequency. Therefore, the effect of the deeper tissues' properties and geometries are not as strong as they are for lower frequencies. With regards to standards, SAR was the valid dose up to 3.0~GHz \cite{IEEEC951}. In the latest version, which was recently published, the upper frequency limit for local SAR validity was set to 6.0 GHz \cite{IEEEC9512019}, being harmonized with ICNIRP guidelines \cite{ICNIRP2019}. For frequencies higher than 3.0~GHz, most EM energy will be deposited in shallow tissues such as skin, muscle and fat, etc., where detailed anatomical information of deep tissues are of low influence. With high-resolution MRI images, the application of the proposed method can be easily extended to these higher frequencies, when required. 

\begin{figure}
\centering
\includegraphics[width=\textwidth]{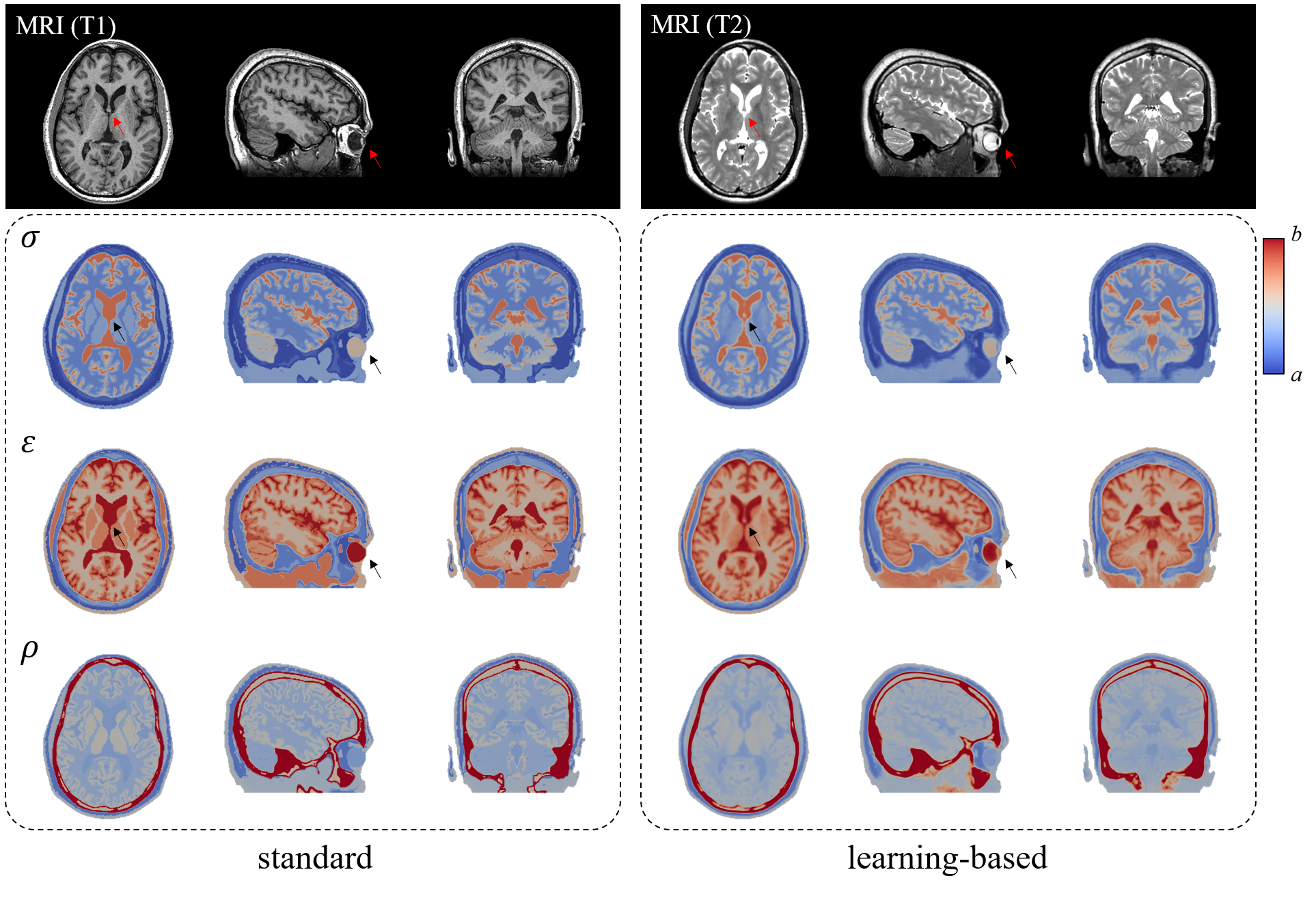}
\caption{Sample result of a subject (case01039). Anatomical images (top) and corresponding physical properties (bottom) computed using the standard and learning-based approaches at 900 MHz. The color scale is defined as [$a,b$] = [0.0, 3.0], [1.0, 70.0], and [800, 1500] for $\sigma$, $\varepsilon$, and $\rho$, respectively. The arrows indicate the regions where the learning-based maps exhibit a higher consistency with the anatomical images.}
\label{fig04}
\end{figure}


\subsection{Metrics for evaluation}

For quantitative evaluation of the SAR distribution, we compute the absolute error using the following equation:

\begin{equation}
E=\frac{\sum_{i \in \Omega} |SAR(i) -  \overline{SAR}(i)|}{\sum_{i \in \Omega} 1}
\end{equation}
where $SAR$ and $\overline{SAR}$ are the SAR maps computed using the standard and learning-based approaches, respectively, and $\Omega$ is the subject head volume. The error values computed at different frequencies are listed in Table~\ref{tab03}. Moreover, the SAR value, which is averaged over 10 g of tissue in a cubic shape (SAR$^{10g}$) excluding the air voxels, of the different subjects is calculated, and the peak spatial averaged SAR ($ps$SAR) values are listed in Table~\ref{tab04}. The relative error is defined as

\begin{equation}
E_{\max}=\frac{|psSAR - \overline{psSAR}|}{psSAR},
\end{equation}

\begin{figure}
\centering
\includegraphics[width=\textwidth]{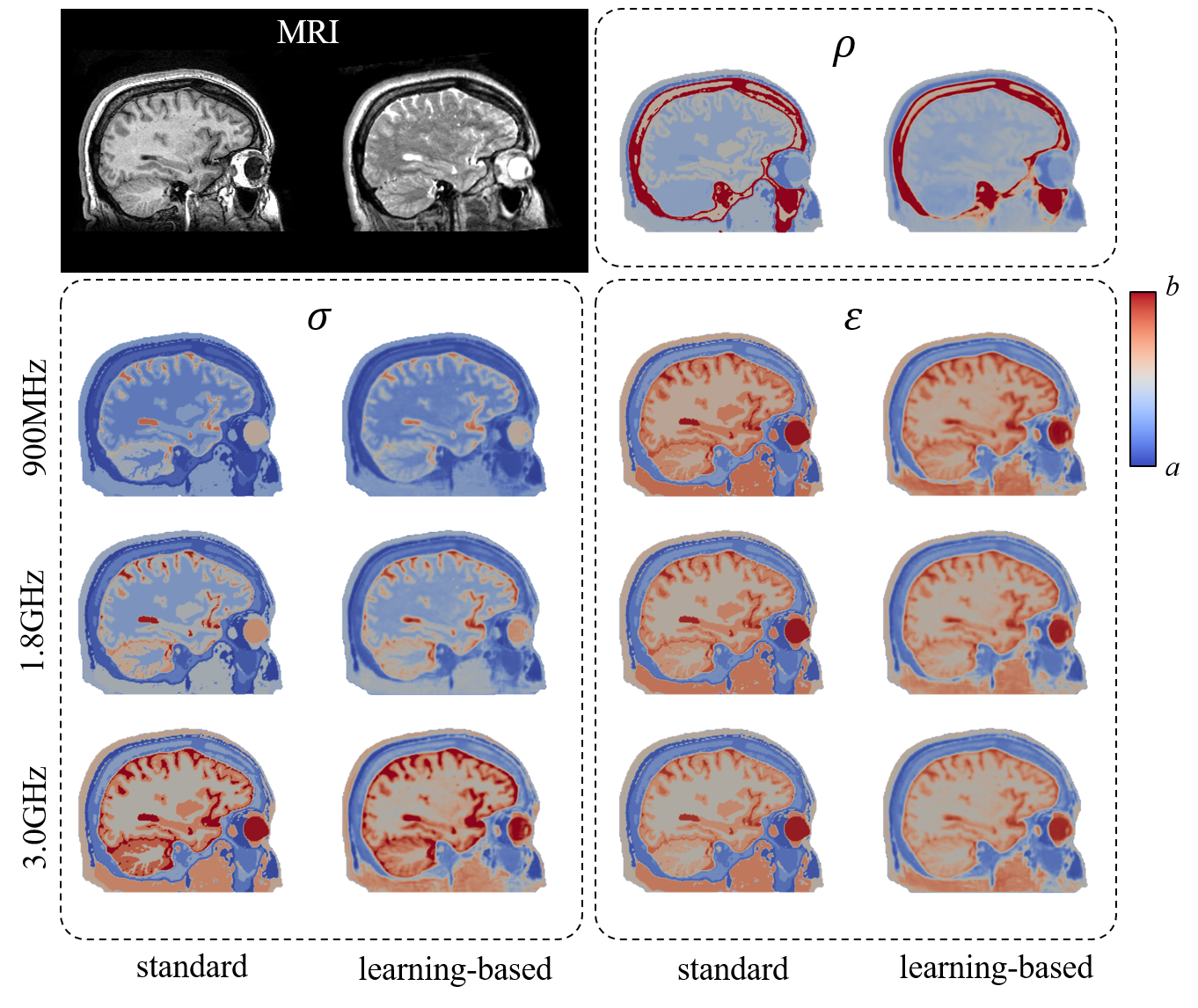}
\caption{Physical properties computed using the standard and proposed (learning-based) approaches at different radio-frequencies. The top left shows the anatomical images. The top right shows the density maps (kg/m$^3$). The bottom left and right represent the conductivity (S/m) and relative permittivity maps at 900 MHz, 1.8 GHz, and 3.0 GHz frequencies, respectively. The color scale is defined as [$a, b$] = [0.0, 3.0], [1.0, 70.0], and [800, 1500] for $\sigma$, $\varepsilon$, and $\rho$, respectively.}
\label{fig05}
\end{figure}


\section{Results}


\subsection{Conductivity, permittivity, and density estimation}

The proposed architecture, shown in figure~\ref{fig02}, is trained to estimate the $\sigma$, $\varepsilon$, and $\rho$ maps using normalized T1- and T2-weighted MRI. Ten arbitrarily selected subjects are employed for optimization of the network parameters through training that uses the cross-validation loss function. In the training phase, two-dimensional MRI slices are mapped to the corresponding uniform physical properties computed using the standard approach shown in figure~\ref{fig01}. The slices are extracted from different orientations (i.e., axial, sagittal, and coronal) in which each slice is used to train the individual network. Training is conducted through 50 epochs with a batch size of four slices and using the ADAM optimizer~\cite{Kingma2014arXiv}. The remaining eight subjects are tapped for evaluation of the training-based estimation. The network outputs from different slicing directions are averaged to generate the estimated $\sigma$, $\varepsilon$, and $\rho$ maps. The learning-based architecture is implemented using four 3.60-GHz Intel (R) Xeon CPU workstation with a 128-GB memory and three NVIDIA GeForce GTX 1080 GPUs. The computation is performed using the GPUs, and a single training phase is completed in approximately 19.2 min. Testing of a single subject requires approximately 12.4 s. An example of the estimated physical properties at a frequency of 900 MHz is shown in figure~\ref{fig04}. From these results, we can clearly see that the learning-based $\sigma$, $\varepsilon$, and $\rho$ maps are highly consistent with those computed using the standard pipeline. However, the learning-based maps show a smoother pattern (see transition within region borders).  

\begin{figure}
\centering
\includegraphics[width=\textwidth]{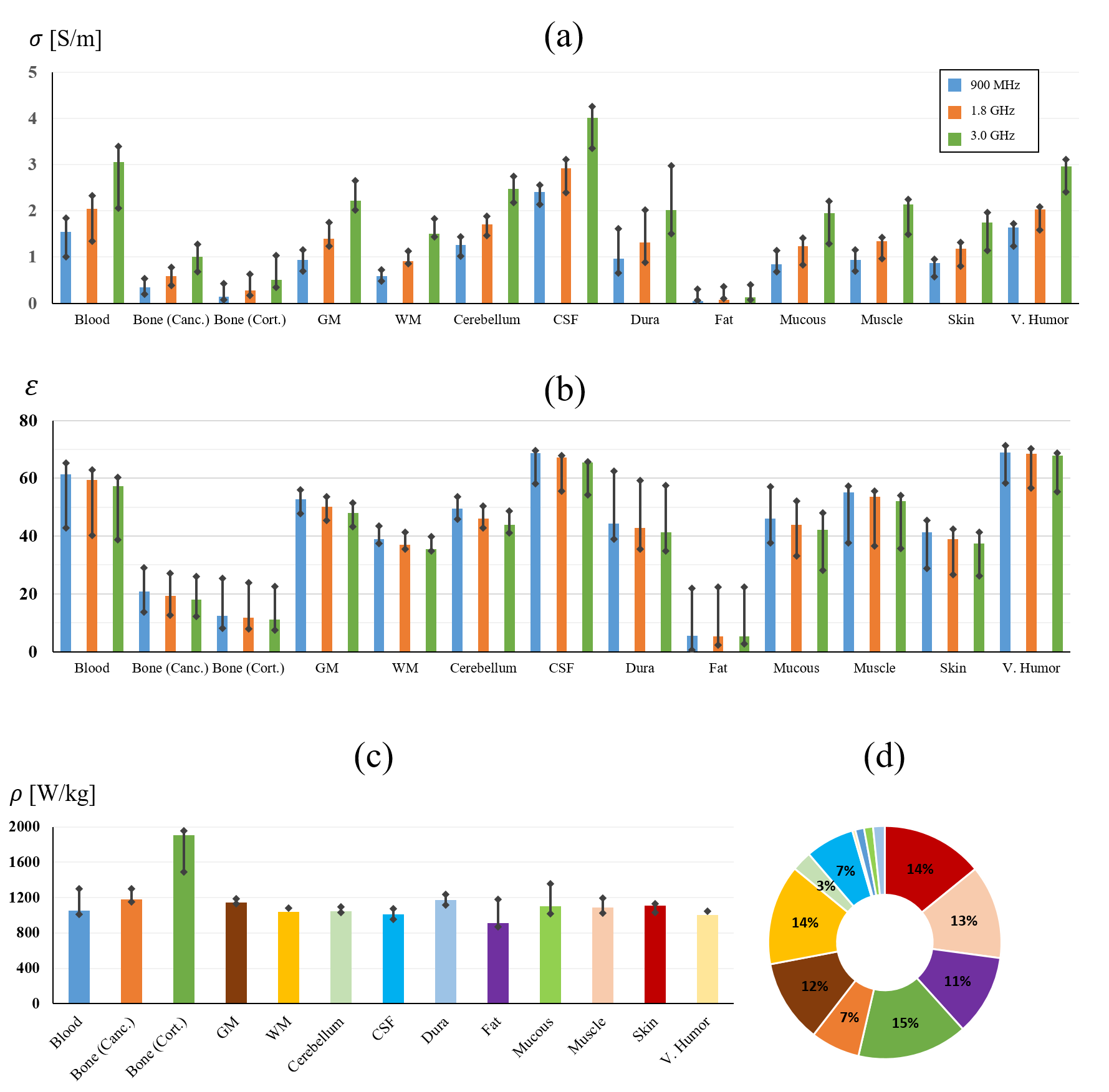}
\caption{Standard and learning-based dielectric properties and tissue-density values of the different head tissues in eight subjects computed at different frequencies. (a) Electrical conductivity, (b) relative permittivity, and (c) tissue density. The color bars represent the constant standard values (Table~\ref{tab01}), and the black lines indicate the learning-based value range (mean$\pm$std). The doughnut chart in (d) shows the average volume ratio of each tissue. Tissue annotation in the standard approach is used as the golden-truth segmentation.}
\label{fig06}
\end{figure}

The training is repeated at frequencies of 1.8 and 3.0 GHz, and an example of the estimated $\sigma$, $\varepsilon$, and $\rho$ maps is shown in figure~\ref{fig05}. From these data, we can observe that the standard method provides a consistent value within the same anatomical regions with high-contrast edges. However, the learning-based values have more realistic (smooth transit) patterns. Moreover, we can observe some regions where the learning-based maps demonstrate higher consistency with the anatomical regions, as shown by the arrows in figure~\ref{fig04}. To highlight how accurate are the estimated learning-based physical-property maps, we compute the mean and standard deviation of each head tissue by testing eight subjects. The values that represent the distributions of $\hat{\sigma}$, $\hat{\varepsilon}$, and $\hat{\rho}$ are shown in figure~\ref{fig06}, which shows that the head tissues presented in high contrast (e.g., white matter, gray matter, and bone) in the anatomical images are estimated with higher accuracy than the low-contrast tissues (e.g., dura, blood, and mucous tissues). 

\begin{figure}
\centering
\includegraphics[width=\textwidth]{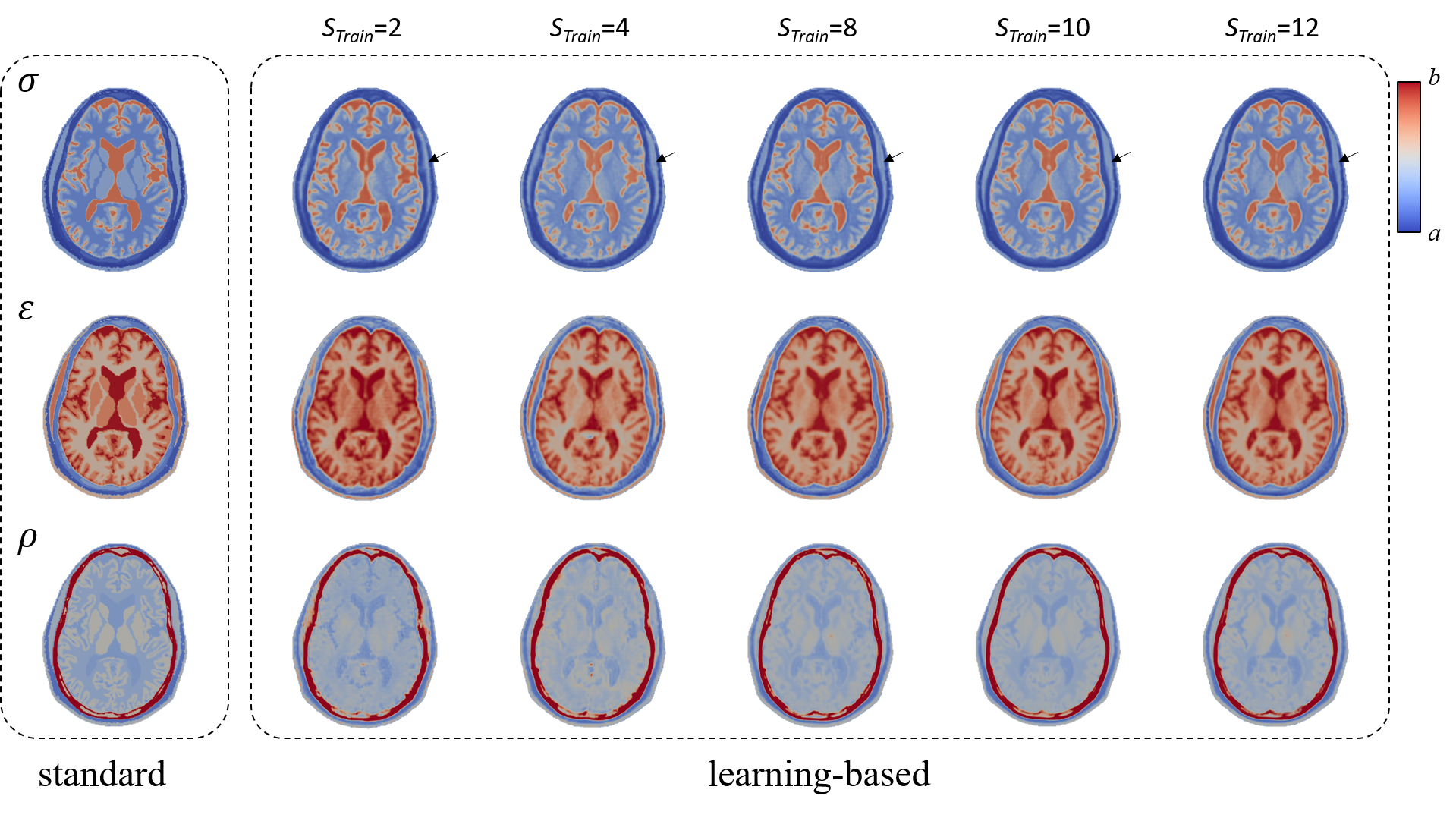}
\caption{Axial slices represent standard and learning-based dielectric properties and tissue-density values for subject (case01039). The learning-based data are computed using network trained with different number of subjects ($S_{Train}$). The color scale is defined as [$a,b$] = [0.0, 3.0], [1.0, 70.0], and [800, 1500] for $\sigma$, $\varepsilon$, and $\rho$, respectively. Arrows are pointing to temporal muscle, where the estimated values can clearly indicate the influence of the number of subjects in training dataset.}
\label{fig07}
\end{figure}

\begin{figure}
\centering
\includegraphics[width=\textwidth]{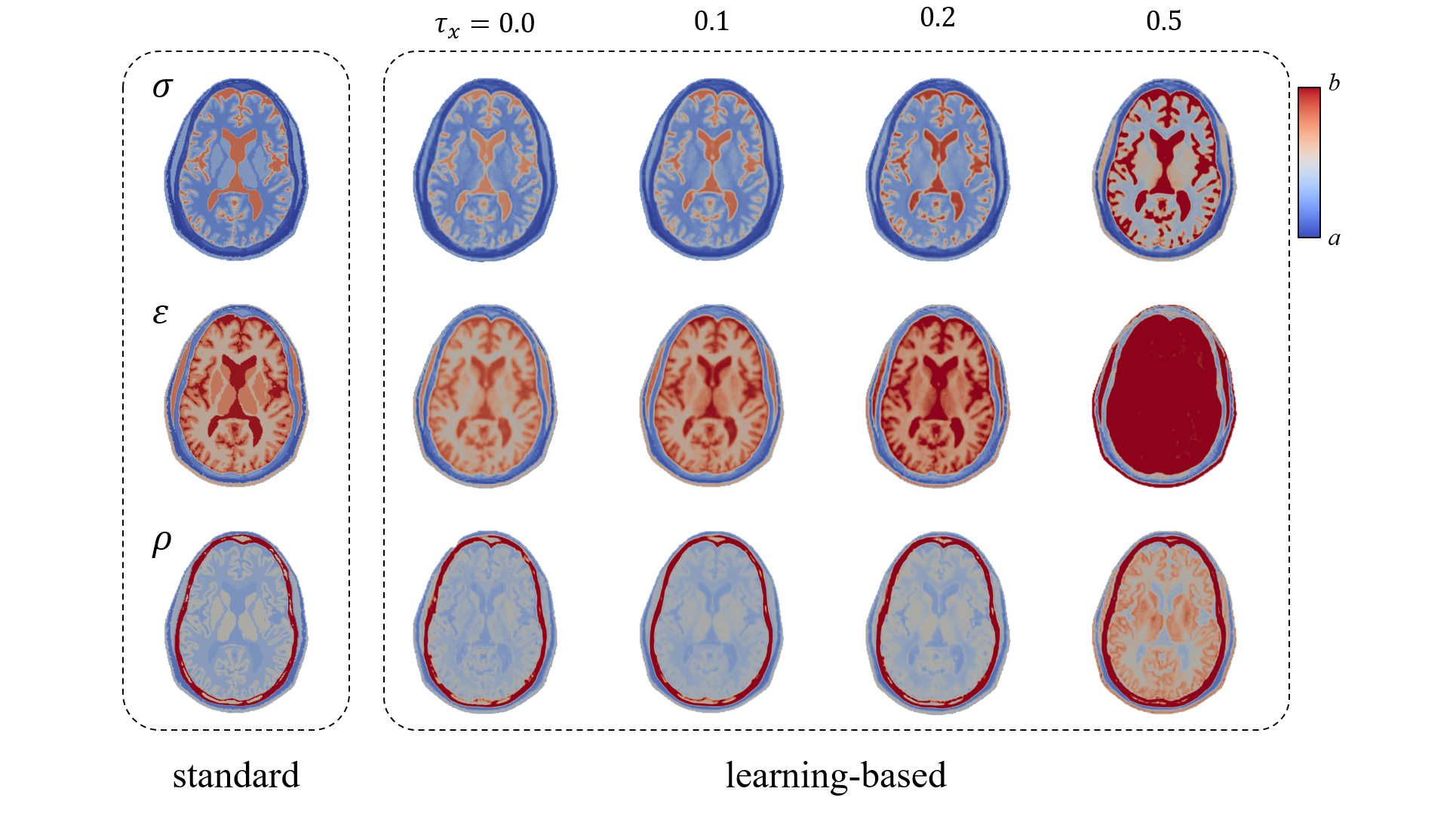}
\caption{Axial slices represent standard and learning-based dielectric properties and tissue-density values for subject (case01039). The learning-based data are computed using different values of parameters $\tau_x$ (i.e. 0.0, 0.1, 0.2, and 0.5), $x\equiv \sigma$, $\varepsilon$, and $\rho$. The color scale is defined as [$a,b$] = [0.0, 3.0], [1.0, 70.0], and [800, 1500] for $\sigma$, $\varepsilon$, and $\rho$, respectively.}
\label{fig08}
\end{figure}


\subsection{Parameter's optimization}

To highlight how the number of training dataset can influence the estimated dielectric properties, the above detailed experiment is repeated with variable number of training subjects (i.e. 2, 4, 8, 10, and 12). Comparison results shown in figure~\ref{fig07} demonstrate that stable estimation can be observed through training using approximately 10 subjects. It is worth noting that this is an arbitrary estimate that is highly related to this specific dataset and might be different when different images are used based on image quality and other related physical properties of imaging acquisition system. It is also interesting to validate the effect of scaling parameters $\tau_{\sigma}$, $\tau_{\varepsilon}$, and $\tau_{\rho}$ in equation~(\ref{eq01}). In general, when $\tau_x=0.0$ ($x \equiv \sigma$, $\varepsilon$, and $\rho$) the training data is normalized such that the maximum value of dielectric properties or tissues density used in training will be set as the upper bound for network prediction values and any value exceed the maximum will be automatically truncated by the sigmoid layers. Therefore, we use $\tau_x=0.1$ to allow network estimation within a frame of approximately 10\% over the maximum trained values. To demonstrate how parameter $\tau_x$ would influence the network output, we have repeated the experiment shown in figure~\ref{fig04} with different $\tau_x$ values and results are shown in figure~\ref{fig08}. It is clear that with $\tau_x=0.0$, network estimate present under-estimate values. On contrary, with higher $\tau_x$ values, network provide over-estimate values, with $0.1$ demonstrate a reasonable value for scaling parameters $\tau_x$. This observation presented in all network outputs consistently.


\subsection{SAR evaluation}

\begin{figure}
\centering
\includegraphics[width=\textwidth]{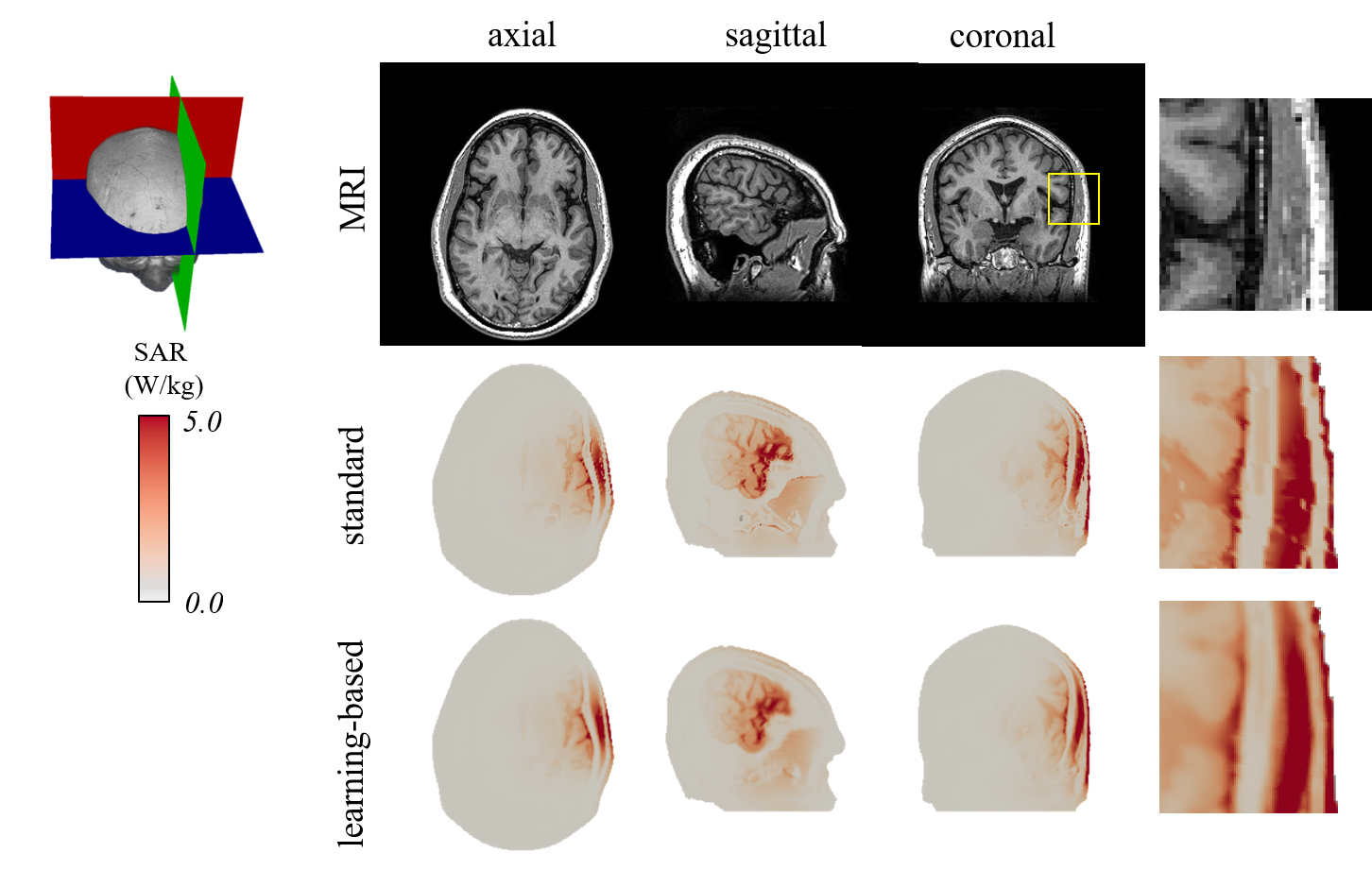}
\caption{SAR maps of one subject at different orientations computed using the standard and learning-based approaches with the corresponding anatomical image for frequency of 900 MHz. The regions labeled by the yellow square are magnified and shown at the rightmost column. At the left side, the axial, sagittal, and coronal-slice positions are shown in blue, green, and red colors, respectively.}
\label{fig09}
\end{figure}

\begin{figure}
\centering
\includegraphics[width=\textwidth]{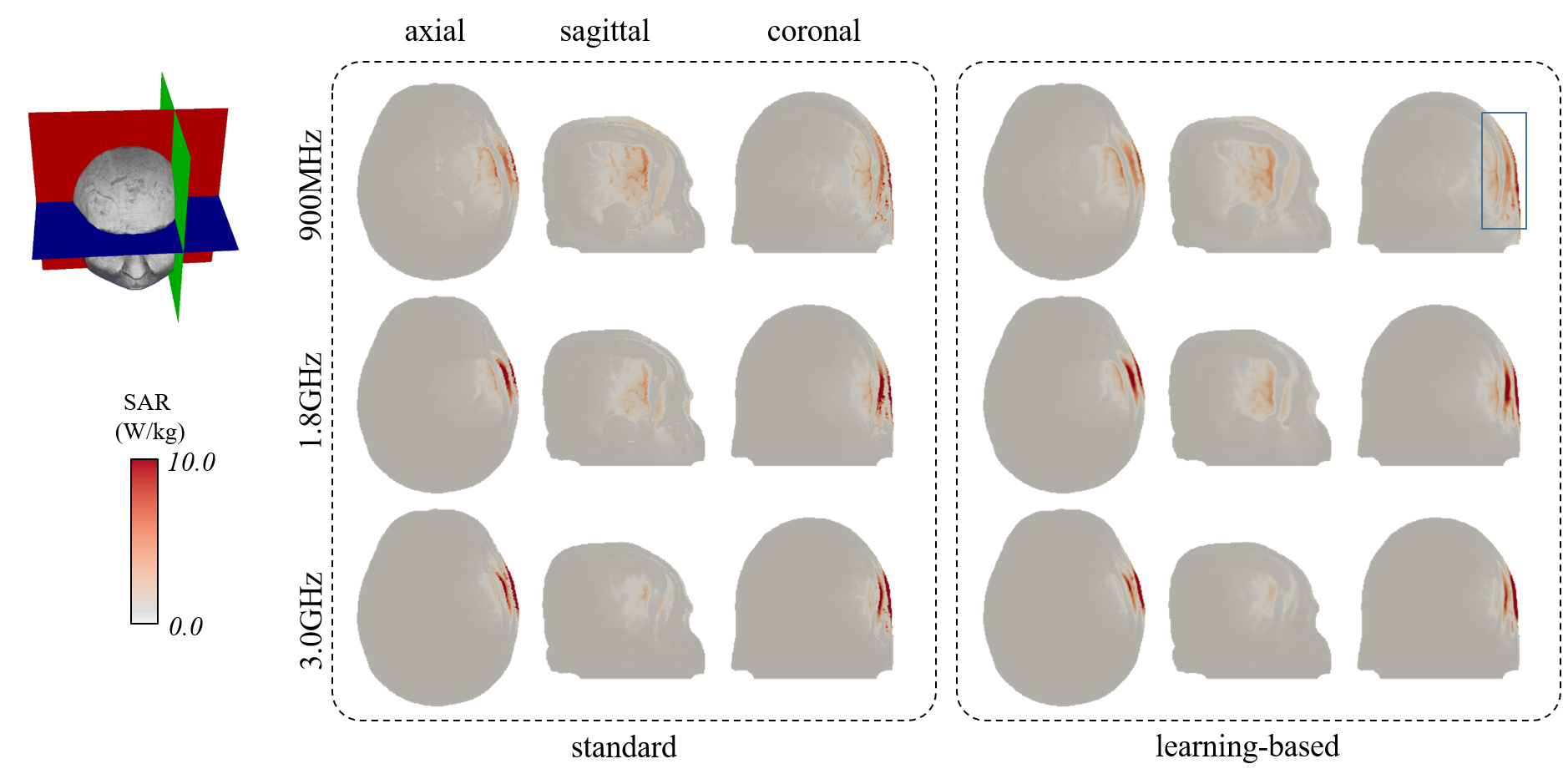}
\caption{SAR distribution at different orientations computed using the uniform and learning-based physical properties at different frequencies. The slice positions are shown at the left side. A magnification of the region labeled with the blue rectangle is shown in figure~\ref{fig11} for different exposure setups.}
\label{fig10}
\end{figure}

\begin{figure}
\centering
\includegraphics[width=\textwidth]{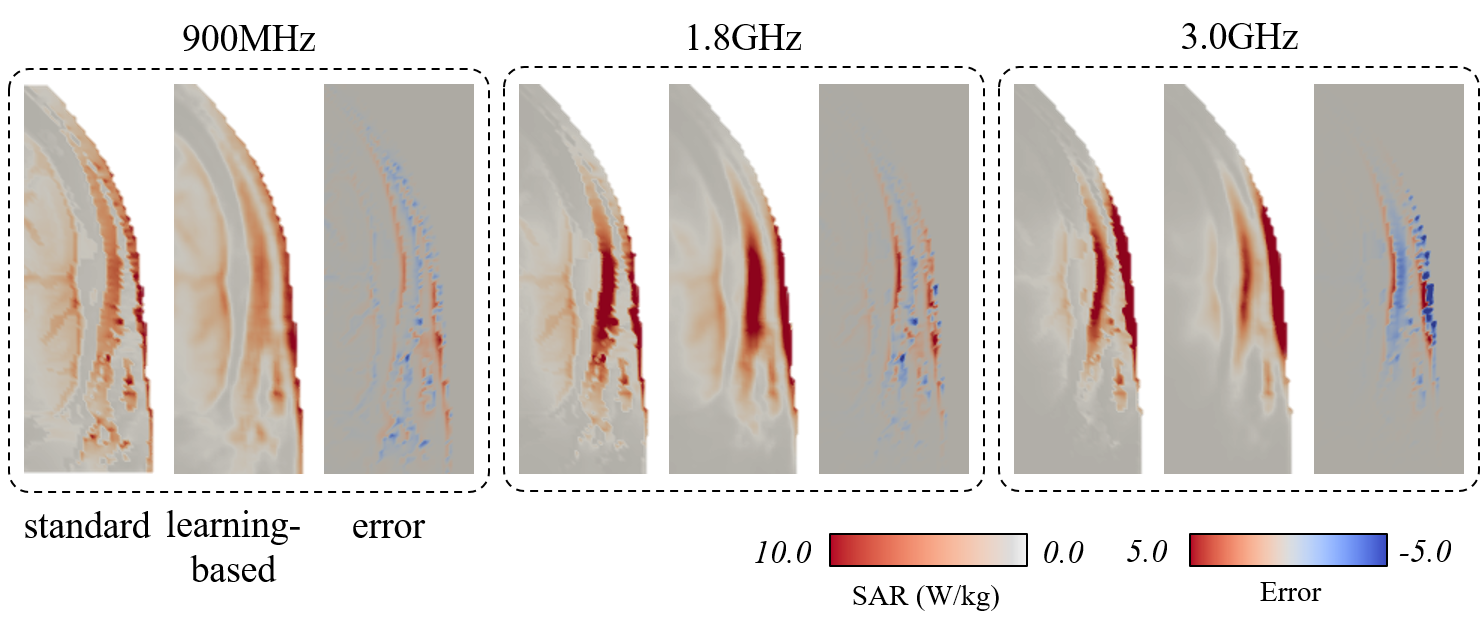}
\caption{Magnified region of the SAR distribution in the coronal direction of the subject shown in figure~\ref{fig10} computed using the different approaches and at different frequencies with their associated error.}
\label{fig11}
\end{figure}

\begin{figure}
\centering
\includegraphics[width=\textwidth]{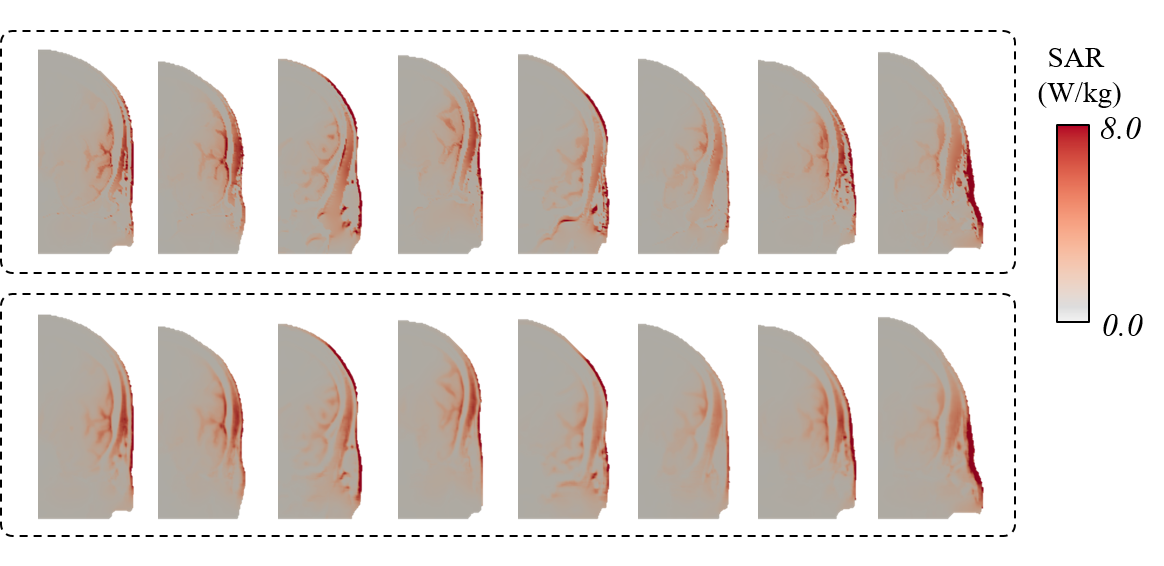}
\caption{Truncated coronal slices of the computed SAR maps of the eight subjects computed using the standard (top) and learning-based (bottom) approaches.}
\label{fig12}
\end{figure}

To evaluate the applicability of the proposed method to radio-frequency dosimetry, two studies are conducted to compute the SAR distributions in the head using the standard physical properties assigned to the segmented models and the learning-based values without segmentation. The radio-frequency exposure from the dipole antenna located close to the left of the temporal lobe, as shown in figure~\ref{fig03}, is considered. The computed results for frequency of 900~MHz are shown in figure~\ref{fig09}, which indicate that consistent results can be obtained in the axial, sagittal, and coronal directions. The major difference is that the SAR distributions computed using the learning-based approach exhibit a texture with smooth transitions. Another example that demonstrates the SAR distribution at different frequencies (i.e. 900~MHz, 1.8~GHz, and 3.0~GHz) is shown in figure~\ref{fig10}. Again, a consistent SAR distributions are observed, as shown in the magnified regions in figure~\ref{fig11}. A single coronal slice of the SAR distribution at 900~MHz in all subjects is extracted and shown in figure~\ref{fig12}. The same behavior with different levels of consistency is observed in all eight tested subjects, which demonstrates the validity of the learning-based approach.

From the list in Table~\ref{tab03}, we can observe a small difference in the SAR values in all subjects with maximum absolute error $E=0.071$, whereas the maximum average error in all subjects is $0.051$. In addition, the quantitative evaluation of the $ps$SAR values (Table~\ref{tab04}) indicates a relatively small change in the different subjects with a maximum error of approximately 17\%. The average maximum 10-g SAR in the eight subjects reflects high matching with errors of 1.65\%, 3.13\%, and 4.95\% at 900 MHz, 1.8 GHz, and 3.0 GHz, respectively. 

\section{Discussion}

In recent studies, deep-learning approaches were applied to directly estimate the induced physical quantities from a given source~\cite{Yokota2019BS,Meliado2019MRM}. \citeasnoun{Meliado2019MRM} proposed a SAR-estimation method using B1$^+$ maps and demonstrated the results using the prostate-imaging application. The results indicated good matching in the spatial-average SAR estimation with reduction in the computation time. However, direct estimation of the averaged SAR from CNN requires careful consideration of several parameters (such as the source position, orientation, and frequency). \citeasnoun{Yokota2019BS} estimated the induced electric field from single kind of coil located at different places.  Thus, these two approaches are powerful only when the source parameters are fixed or its flexibility is small because the variations of all these potential variables require a huge amount of training dataset that demonstrate several potential challenges.

\begin{table}
\centering
\footnotesize
\caption{Absolute error ($E$) of SAR values for different subjects at different frequencies ($\times 10^{-2}$).}
\setlength{\tabcolsep}{3pt}
\begin{tabular}{ c|l |c  |  c|  c}
\hline
\#&{\bf Subject}	  &900 MHz  & 1.8 GHz &  3.0 GHz	\\
\hline
\hline
~1&case01017 & 4.536 & 3.200 & 2.441 \\
~2&case01019 & 4.878 & 3.033 & 2.352 \\
~3&case01025 & 7.114 & 4.751 & 3.564 \\
~4&case01028 & 4.278 & 2.628 & 1.949 \\
~5&case01034 & 5.909 & 3.473 & 1.909 \\
~6&case01039 & 4.443 & 3.357 & 2.878 \\
~7&case01042 & 5.260 & 3.701 & 3.459 \\
~8&case01045 & 4.075 & 2.944 & 1.686 \\
\hline \hline
& average & 5.062 & 3.386 & 2.529 \\
\hline
\end{tabular}
\label{tab03}
\end{table}

\begin{table}
\centering
\footnotesize
\caption{$ps$SAR computed from SAR maps generated using standard (S) and learning-based (L) approaches with relative error ($E_{\max}$) for different subjects at different frequencies.}
\setlength{\tabcolsep}{3pt}
\begin{tabular}{ c|l |c |c |c  |  c| c |c |  c| c|c}
\hline
\multirow{2}{*}{\#}&\multirow{2}{*}{\bf Subject}	  &\multicolumn{3}{c|}{ 900 MHz}  & \multicolumn{3}{c|}{ 1.8 GHz} & \multicolumn{3}{c}{ 3.0 GHz}	\\
\cline{3-11} 
&&S & L & E[\%]&S & L& E[\%]& S & L& E[\%]\\
\hline
\hline
~1&case01017 & 2.780 & 2.945 & ~6.33 & 3.610 & 3.729 & ~3.29 & 3.541 & 2.947 & 16.77 \\
~2&case01019 & 3.335 & 3.255 & ~2.40 & 3.185 & 3.205 & ~0.65 & 3.486 & 3.630 & ~4.14 \\
~3&case01025 & 2.809 & 2.946 & ~4.88 & 3.093 & 3.374 & ~9.10 & 3.896 & 3.757 & ~3.57 \\
~4&case01028 & 3.047 & 3.156 & ~3.56 & 2.815 & 2.856 & ~1.42 & 3.306 & 3.322 & ~0.48 \\
~5&case01034 & 3.277 & 3.185 & ~2.81 & 3.142 & 3.146 & ~0.12 & 2.698 & 2.867 & ~6.24 \\
~6&case01039 & 2.464 & 2.638 & ~7.03 & 3.271 & 3.604 & 10.19 & 3.225 & 2.745 & 14.89 \\
~7&case01042 & 2.614 & 2.729 & ~4.38 & 3.142 & 3.444 & ~9.59 & 2.723 & 2.289 & 15.95 \\
~8&case01045 & 3.627 & 3.485 & ~3.91 & 5.027 & 4.781 & ~4.88 & 2.720 & 2.773 & ~1.92 \\
\hline \hline
&average & 2.993 & 3.042 & ~1.65 & 3.411 & 3.517 & ~3.13 & 3.200 & 3.041 & ~4.95 \\
\hline
\end{tabular}
\label{tab04}
\end{table}

The use of personalized electromagnetic exposure requires enhancement of the current standard approach. However, appropriate calculations require a high-quality annotation of the different tissue compositions in the biological object under study. This process is considered to be a challenging task considering the large variability in the electrical field in a human body and resulting SAR distributions. This work presents a novel framework for automatic estimation of the dielectric properties and tissue density of all head tissues directly from MRI. The proposed method is straightforward to use and does not require highly sophisticated imaging protocols. The learning-based dielectric properties can be estimated with high accuracy without segmentation. Moreover, the tissue mass density is estimated using a single-step network validation. High consistency is observed with the uniform values used in the training dataset. This result can be attributed to the ability of the learning architecture to estimate the water contents using the gray-scale value of anatomical images. The learning-based estimated values are used to assess the SAR distribution, and the results provide good matching with those computed using the standard approach. Nevertheless, the learning-based approach provides a smoother distribution with a relatively small average difference. For further validation as well as for practical application, the peak value of 10-g average SAR is computed. The results indicate that the learning-based method using CondNet has a high-consistency value with the conventional approach. It achieves an average relative difference of less than 4.95\%, which is well within the range of computational uncertainty attributable to different electromagnetic computations~\cite{Beard2006TEMC}.

A substantial number of computational data are needed for variability analysis, namely, different exposure scenarios with different sources, distances, and orientation, because the antenna-human coupling is significant, especially when the antenna-head separation is in the near-field regime, where the mutual coupling is significant. Therefore, the generation of a personalized head model is one way to conduct variability analysis for human safety application at radio-frequency. In the low-frequency dosimetry for medical application, more than 200,000 cases are considered to learn the different source positions~\cite{Yokota2019BS}. Developing a learning-based method that can consider a wide range of exposure scenarios with all potential varieties of experimental setup remains a challenge. However, the time can be significantly reduced by avoiding extensive segmentation, especially when the number of tissues is large and with some tissues hardly being observed in the anatomical image. This work is a step forward toward a fully automatic simulation using deep learning.


\section{Conclusion}

Learning-based estimation of the physical properties of human head models has been presented. Once the network is trained, the estimated non-uniform values of conductivity, permittivity, and mass-density values can be automatically generated in a single shot. The learning-based dielectric properties as well as the tissue density demonstrate high-consistency patterns with those generated using the standard pipeline that employs time-consuming segmentation. The proposed approach demonstrates several advantages. It provides a more robust approach with better representation of the corresponding anatomy. The heterogeneous physical properties exhibit a smoother pattern, resulting in a smoother SAR distribution. The computed results for a dipole antenna indicate that the SAR distribution of the proposed approach is highly consistent with that of the standard approach. 


\section*{Acknowledgment}
This work was supported in part by JSPS KAKENHI Grant Number 17H00869.



{\color{white}\cite{Gabriel1996PMBII}.}
\section*{References}
\bibliographystyle{dcu}
\bibliography{Ref1}
\end{document}